\documentclass[10pt,twocolumn,letterpaper]{article}

\usepackage[pagenumbers]{cvpr} 

\definecolor{cvprblue}{rgb}{0.21,0.49,0.74}
\usepackage[pagebackref,breaklinks,colorlinks,allcolors=cvprblue]{hyperref}
\usepackage[accsupp]{axessibility}
\usepackage{amsmath}
\usepackage{float}
\usepackage{adjustbox}
\usepackage{multirow}
\usepackage[table]{xcolor}

\newcommand{\blockcomment}[1]{}
\makeatletter
\def\blfootnote{\xdef\@thefnmark{}\@footnotetext}
\makeatother

\def\paperID{***} 
\def\confName{CVPR}
\def\confYear{2026}

\title{Cubic Discrete Diffusion: Discrete Visual Generation on High-Dimensional Representation Tokens}

\def\Ours{{CubiD}\xspace}
\usepackage{wrapfig}

\author{
  \textbf{Yuqing Wang}\textsuperscript{1} 
  \quad \textbf{Chuofan Ma}\textsuperscript{1} 
  \quad \textbf{Zhijie Lin}\textsuperscript{2$\dagger$} 
  \quad \textbf{Yao Teng}\textsuperscript{1} 
  \quad \textbf{Lijun Yu}\textsuperscript{3} \\
  \quad \textbf{Shuai Wang}\textsuperscript{4} 
  \quad \textbf{Jiaming Han}\textsuperscript{5} 
  \quad \textbf{Jiashi Feng}\textsuperscript{2}  
  \quad \textbf{Yi Jiang}\textsuperscript{2} 
  \quad \textbf{Xihui Liu}\textsuperscript{1}\footnotemark[1]  \\
\textsuperscript{1}University of Hong Kong  \quad
\textsuperscript{2}ByteDance Seed \quad
\textsuperscript{3}Carnegie Mellon University\quad \\
\textsuperscript{4}Nanjing University\quad
\textsuperscript{5}The Chinese University of Hong Kong 
}

\begin{document}
\maketitle
\blfootnote{$\dagger$Project lead. \quad $*$Corresponding author.}
\begin{abstract}
Visual generation with discrete tokens has gained significant attention as it enables a unified token prediction paradigm shared with language models, promising seamless multimodal architectures. However, current discrete generation methods remain limited to low-dimensional latent tokens (typically 8-32 dims), sacrificing the semantic richness essential for understanding. While high-dimensional pretrained representations (768-1024 dims) could bridge this gap, their discrete generation poses fundamental challenges.
In this paper, we present Cubic Discrete Diffusion (CubiD), the first discrete generation model for high-dimensional representations. 
CubiD performs fine-grained masking throughout the high-dimensional discrete representation—any dimension at any position can be masked and predicted from partial observations. This enables the model to learn rich correlations both within and across spatial positions, with the number of generation steps fixed at $T$ regardless of feature dimensionality, where $T \ll hwd$. On ImageNet-256, CubiD achieves state-of-the-art discrete generation with strong scaling behavior from 900M to 3.7B parameters. Crucially, we validate that these discretized tokens preserve original representation capabilities, demonstrating that the same discrete tokens can effectively serve both understanding and generation tasks. We hope this work will inspire future research toward unified multimodal architectures.
Code is available at: \url{https://github.com/YuqingWang1029/CubiD}. 
\end{abstract}
    
\section{Introduction}
\label{sec:intro}

\begin{figure}[!t]
\centering
\includegraphics[width=\linewidth]{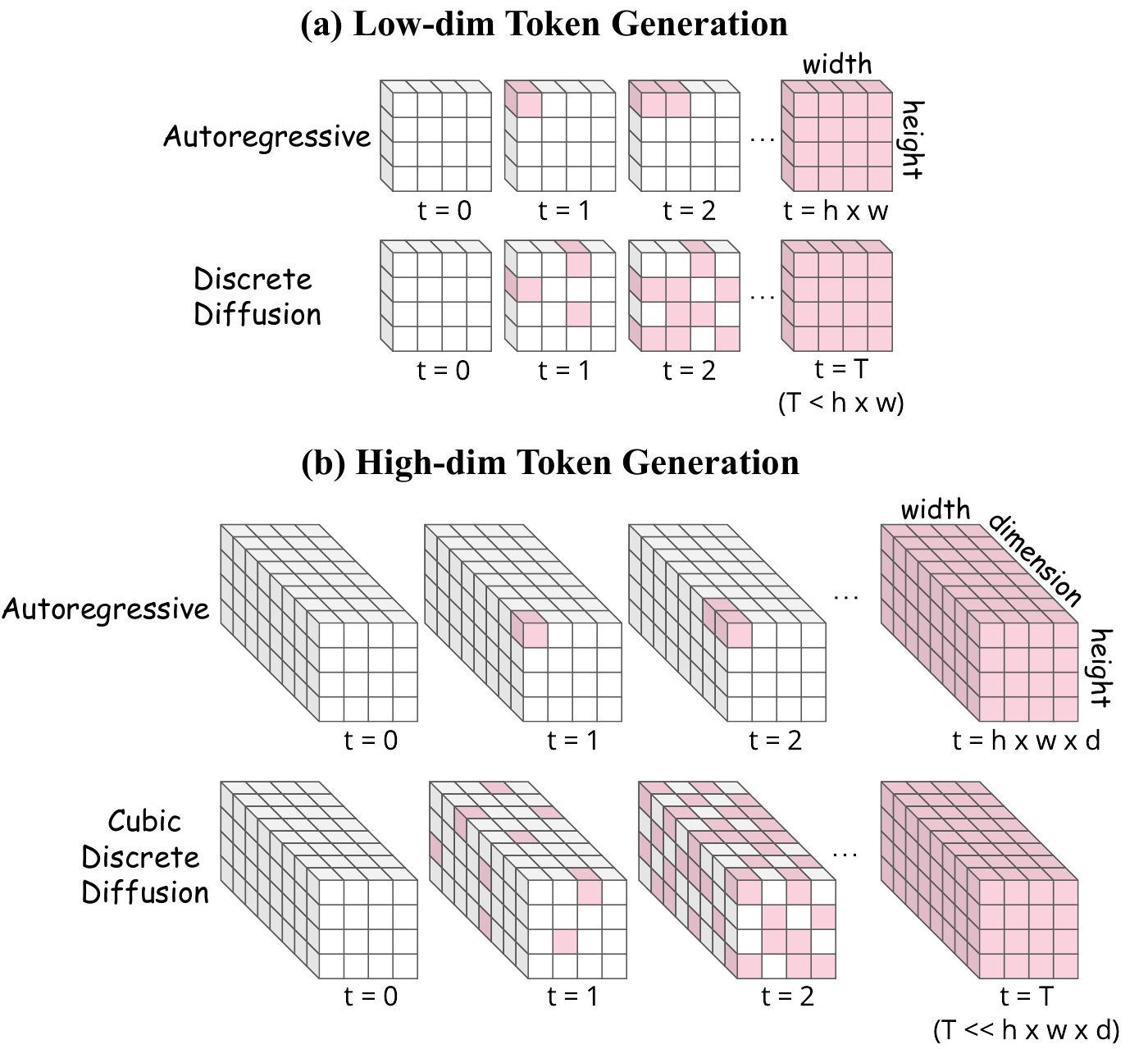}
\vspace{-2em} 
\caption{\textbf{Comparison of discrete visual generation approaches.} 
(a) \textbf{Low-dimensional token generation:} Both methods operate at the spatial level—autoregressive requires $h \times w$ sequential steps, while discrete diffusion achieves parallel generation in $T < h \times w$ iterations. 
(b) \textbf{High-dimensional token generation:} Autoregressive becomes intractable ($h \times w \times d$ steps), and standard discrete diffusion cannot model intra-position dependencies. Our Cubic Discrete Diffusion performs fine-grained masking across the entire 3D tensor—any dimension at any position can be masked and predicted independently—enabling efficient generation in $T \ll h \times w \times d$ iterations while capturing both spatial and dimensional correlations.}
\vspace{-15pt}
\label{fig:teaser}
\end{figure}

\begin{figure*}[!t]
\centering
\includegraphics[width=\linewidth]{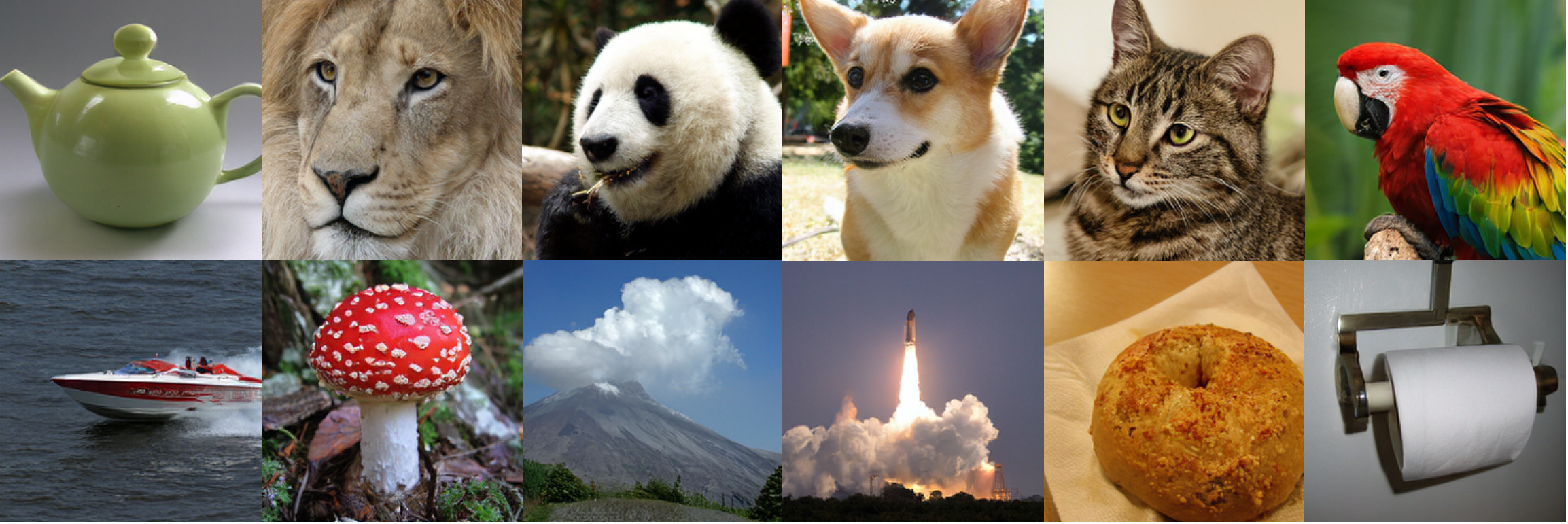}
\vspace{-18pt}
\caption{\textbf{Generated samples from CubiD.} Class-conditional generation results on ImageNet 256×256 using high-dimensional representation tokens from DINOv2-B encoder, demonstrating fine details and textures across diverse categories.}
\vspace{-15pt}
\label{fig:vis}
\end{figure*}

The pursuit of unified multimodal modeling~\cite{team2024chameleon,cui2025emu3,Xie2024Showo} requires both language and vision to operate on semantically meaningful tokens. While language models have long benefited from semantic tokens that naturally support both understanding and generation, visual models remain fragmented—using high-dimensional semantic features for understanding but low-dimensional compressed tokens~\cite{vae,vqvae,esser2020taming,vit-vqgan,Zheng_2023_CVQ} for generation.  
 Recent advances~\cite{emu2,blip3o, rae} have shown that high-dimensional representation features (768-1024 dimensions) can achieve high-quality reconstruction, offering a path forward. 
For discrete generative models~\cite{gpt3,sun2024autoregressive,tian2024visual}, which share the token-based paradigm with language models, adopting such high-dimensional representation tokens is particularly compelling, as it would allow visual generation to leverage the same semantic richness that has proven essential for understanding, potentially enabling more coherent unified architectures.

However, high-dimensional representations pose significant challenges for discrete generative modeling. The first is how to discretize these features while maintaining their representation quality. Traditional Vector Quantization~\cite{vqvae} methods that work well in low dimensions (8-32) fail at 768-1024 dimensions due to the curse of dimensionality—data points become sparsely distributed, making clustering ineffective, and the codebook size required for adequate coverage grows exponentially. The quantized features inevitably drift from the original representations, corrupting the semantic information essential for understanding. Dimension-wise quantization~\cite{tokenbridge} offers a promising solution. By treating each dimension independently rather than quantizing entire vectors jointly, it sidesteps the clustering problems in high-dimensional spaces. As a training-free method, it can be directly applied to frozen pretrained features, making discretization tractable at 768+ dimensions.
We validate this approach on multimodal understanding tasks: dimension-wise quantized features achieve nearly identical performance to continuous features, while VQ suffers substantial degradation (Table~\ref{tab:understanding}). This result confirms that properly discretized high-dimensional tokens preserve semantic quality for understanding tasks, 
establishing them as viable unified representations.

The more fundamental challenge lies in modeling such high-dimensional discrete tokens. While dimension-wise quantization successfully preserves semantic quality, the resulting representation contains $h \times w \times d$ discrete tokens (196,608 for a typical $16 \times 16 \times 768$ configuration). As illustrated in Figure~\ref{fig:teaser}(b), direct sequential generation requires $O(hwd)$ steps, which is intractable, while standard discrete diffusion methods cannot capture the dependencies across dimensions within each spatial position.
To make this problem tractable, we need a method that avoids sequential bottlenecks while preserving the rich dependency structure across both spatial and dimensional axes. We observe that the $h \times w \times d$ tensor has inherent multi-dimensional structure that can be exploited—rather than treating spatial positions as atomic units or requiring sequential generation of all dimensions, we can break these rigid boundaries and operate flexibly across the entire tensor.  

We propose \textbf{Cubic Discrete Diffusion (CubiD)}, a masked diffusion method~\cite{austin2021structured,chang2022maskgit,lou2023discrete} for high-dimensional discrete generation. Our key insight is to perform fine-grained masking across the three-dimensional $h \times w \times d$ tensor. Unlike existing methods~\cite{chang2022maskgit} that mask entire spatial positions, our approach treats this tensor as a unified cubic space where any subset of dimensions at any position can be masked and predicted from partial observations. This allows the model to learn complex dependencies both within and across spatial locations. As shown in Figure~\ref{fig:teaser}(b), during generation, CubiD starts from a fully masked tensor and iteratively refines it through progressive unmasking, randomly selecting tokens across the entire tensor to unmask at each step until reaching the complete representation.

This approach offers two main advantages. First, it effectively models complex dependencies in high-dimensional tensors—learning both intra-position correlations (how dimensions relate within a spatial location) and inter-position patterns (how features propagate spatially)—through bidirectional attention over partially observed values. Second, it decouples generation complexity from dimensionality: unlike autoregressive methods that scale with $O(hwd)$, our iterative refinement requires a fixed number of steps $T$ regardless of feature dimensionality, benefiting from the semantic redundancy inherent in high-dimensional representations. By transforming an intractable sequential process into hundreds of parallel iterations, CubiD makes high-dimensional discrete generation computationally feasible while maintaining the modeling capacity necessary for high-quality synthesis. 

Extensive experiments validate our approach. We first verify that dimension-wise quantization preserves both understanding and reconstruction capabilities of the original continuous representations. In ablation studies, we compare our fine-grained cubic masking against alternative strategies: treating spatial positions or dimensions as groups significantly degrades performance, confirming the necessity of element-wise masking across the 3D tensor. The method also exhibits strong scaling behavior from 900M to 3.7B parameters and generalizes well across different representation encoders (DINOv2~\cite{oquab2023dinov2} and SigLIP2~\cite{tschannen2025siglip2}). On ImageNet 256×256~\cite{imagenet}, CubiD achieves a competitive 1.88 FID score with 768-dimensional discrete tokens, establishing that high-dimensional discrete generation is both feasible and effective.

Our contributions are summarized as follows:

\begin{itemize}
    \item We demonstrate that proper discretization of high-dimensional representation tokens can preserve their original semantic capabilities, establishing the viability of unified discrete representations for both understanding and generation.
    
    \item We propose Cubic Discrete Diffusion, a novel method that addresses the fundamental modeling challenge of high-dimensional discrete generation by treating the $h \times w \times d$ tensor as a unified space with fine-grained masking, making discrete generative models tractable at high dimensionality.
    
    \item We achieve state-of-the-art discrete generation results on ImageNet 256×256, with strong scaling behavior from 900M to 3B parameters and generalization across different representation encoders, demonstrating the effectiveness of discrete diffusion for high-dimensional visual generation.
\end{itemize}

\section{Related Work}

\paragraph{Visual Tokenization}
Visual tokenization is commonly used to convert images into latent representations that support image reconstruction and generation. In the traditional VAE tokenizers~\cite{vae,dai2019diagnosing}, an encoder first compresses an image into a low-dimensional continuous latent map (typically with 4–32 dimensions) and then a decoder reconstructs the corresponding image with the latent as input. The encoder and decoder of these tokenizers are jointly trained for the reconstruction task. Building on this framework, discrete tokenizers further quantize each vector from the latent maps into one or several tokens~\cite{esser2020taming,yu2023language,tokenbridge,mentzer2023finite,bsq,infinity}, enabling discrete image generation.
More recently, representation-based tokenizers~\cite{rae,zheng2025vision-foundation,shi2025latent} have emerged. Most of these methods use a frozen pretrained vision foundation model~\cite{oquab2023dinov2,tschannen2025siglip2} as the encoder and further train additional adapters to project its outputs into low-dimensional latents. In contrast, RAE~\cite{rae} directly uses high-dimensional DINOv2~\cite{oquab2023dinov2} or SigLIP~\cite{tschannen2025siglip2} features as latents (768+ dimensions) without any adaptation, and a specially designed training schedule is applied to these high-dimensional latents to adapt the continuous diffusion models for generation. In this paper, we first transform high-dimensional features from vision foundation models into discrete tokens and then train generative models on those tokens. 

\paragraph{Discrete Visual Generation}
Discrete visual generation performs image generation based on sequences of discrete tokens. Autoregressive models~\cite{ramesh2021dalle,yu2022scaling,sun2024autoregressive,wang2024loong,kondratyuk2023videopoet,par,liu2024lumina} generate tokens sequentially via the next-token prediction paradigm. Although these models can generate high-quality images, they require $O(N)$ generation steps for $N$ tokens, making this paradigm computationally expensive for high-resolution images.
To improve sampling efficiency, discrete diffusion models~\cite{chang2022maskgit} have been introduced. Instead of generating tokens sequentially, they generate multiple tokens in parallel, thereby achieving higher efficiency.
Like continuous diffusion models, discrete diffusion models also learn to restore corrupted tokens, with corruption defined by absorbing-state~\cite{chang2022maskgit,lou2023discrete,weber2024maskbit,nie2025large}, uniform~\cite{austin2021structured}, or Gaussian-like transitions~\cite{austin2021structured,lou2023discrete}. Among these, the absorbing-state transition is the predominant choice due to its strong empirical performance~\cite{nie2025large}. It corrupts tokens into a special \texttt{[MASK]} state, aligning with representative masked generative models such as BERT~\cite{devlin2019bert} and MaskGIT~\cite{chang2022maskgit}. Existing autoregressive and discrete diffusion models perform well when each image is represented by a small number of discrete tokens derived from low-dimensional latents. However, when representation-based tokenizers produce more tokens per latent, the total token count grows dramatically and existing models become impractical. Therefore, in this work, we extend discrete diffusion models to more efficiently handle tokens derived from high-dimensional latents.

\section{Method}
\label{sec:method}

\begin{figure*}[!tbp]
  \centering
  \includegraphics[width=\linewidth]{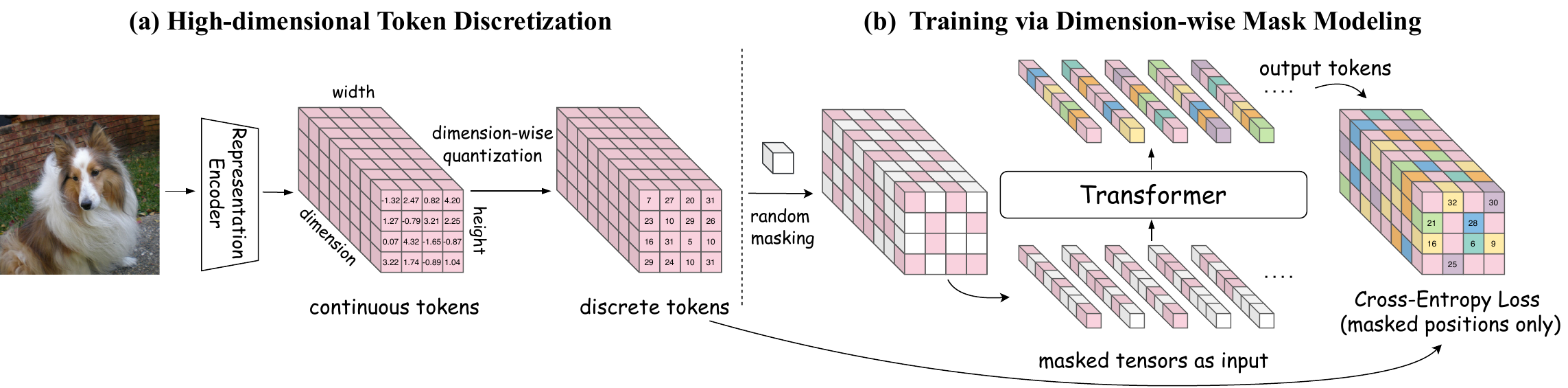}
  \vspace{-15pt}
    \caption{\textbf{Overview of Cubic Discrete Diffusion.} 
(a) \textbf{High-dimensional Token Discretization.} Given an input image, a frozen representation encoder extracts continuous tokens, which are then discretized through dimension-wise quantization into $h \times w \times d$ discrete tokens. 
(b) \textbf{Training via Dimension-wise Mask Modeling.} During training, we randomly mask tokens across both spatial and dimensional axes of the tensor (white: masked tokens, pink: visible ground truth tokens, other colors: predicted tokens). The transformer learns to predict these masked tokens from the unmasked context, capturing the complex dependencies across both spatial and dimensional axes. 
}
\vspace{-1em}
\label{fig:overview}
  \label{fig:train}
\end{figure*}

Our goal is to enable discrete generative modeling of high-dimensional representation tokens from frozen pretrained encoders. This requires two steps: discretizing the continuous high-dimensional features, and modeling the resulting discrete token distribution. We first review the necessary preliminaries: high-dimensional features from pretrained encoders and dimension-wise quantization that enables tractable discretization (Sec.~\ref{sec:prelim}). The core challenge—and our main contribution—lies in modeling the joint distribution of the resulting $h \times w \times d$ discrete tokens, an exponentially large space where traditional methods fail. We propose Cubic Discrete Diffusion (CubiD), which performs masked prediction across both spatial and dimensional axes simultaneously. By masking and predicting at the dimension level, CubiD captures complex inter-dimensional dependencies while enabling efficient parallel generation, transforming intractable sequential modeling into practical iterative refinement (Sec.~\ref{sec:cubiD}).


\subsection{Preliminaries}
\label{sec:prelim}

\noindent\textbf{High-dimensional Representation Tokens.} 
Our method operates on features from frozen pretrained vision encoders. Given an input image $\mathbf{x} \in \mathbb{R}^{H \times W \times 3}$, a pretrained encoder $E$ (e.g., DINOv2~\cite{oquab2023dinov2}, SigLIP2~\cite{tschannen2025siglip2}) with patch size $p$ produces a feature map $\mathbf{z} = E(\mathbf{x}) \in \mathbb{R}^{h \times w \times d}$, where $h = H/p$, $w = W/p$, and $d$ is the feature dimension (typically 768-1024). These encoders produce semantically rich, high-dimensional features that capture both local details and global semantic structures, in contrast to the low-dimensional compressed spaces (8-32 dims) commonly used in generative modeling.

\noindent\textbf{Dimension-wise Quantization.} 
To discretize these high-dimensional features, we adopt dimension-wise quantization~\cite{tokenbridge}, which operates directly on frozen encoder features without any retraining. As shown in Figure~\ref{fig:train}(a), it independently quantizes each continuous value into $L$ discrete levels:
\begin{equation}
    q_{x,y,i} = \text{Quantize}(z_{x,y,i}; L),
\end{equation}
where $z_{x,y,i} \in \mathbf{z}$ denotes the $i$-th dimension at spatial position $(x,y)$, and $\text{Quantize}(\cdot; L)$ maps continuous values to discrete indices in $\{0, ..., L-1\}$. Unlike vector quantization which struggles to cover high-dimensional spaces with fixed-size codebooks, this method treats each dimension independently, making it tractable even for 768-dimensional features. The resulting $h \times w \times d$ discrete tokens maintain their tensor structure.
More details can be found in~\cite{tokenbridge}. Through experiments on understanding tasks, we verify that this discretization preserves the semantic quality of the original representations (Table~\ref{tab:understanding}).

\subsection{Cubic Discrete Diffusion}
\label{sec:cubiD}

\begin{figure*}[!tbp]
  \centering
  \includegraphics[width=\linewidth]{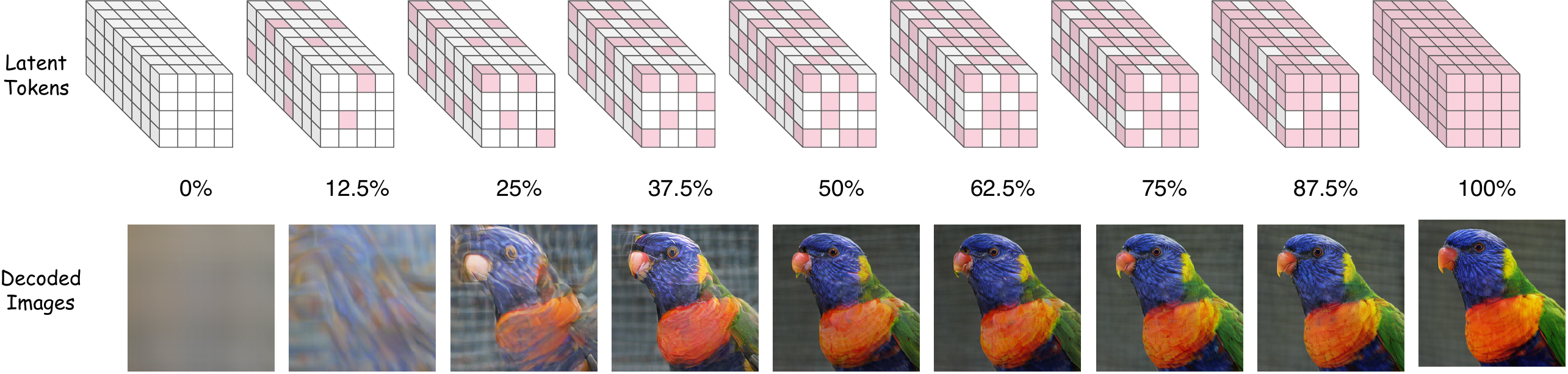}
  \vspace{-15pt}
    \caption{\textbf{Inference process of \Ours.} Top row shows the latent token state (white: masked, pink: unmasked), bottom row shows corresponding decoded images. During generation, \Ours starts from a fully masked tensor (0\%) and progressively unmasks tokens until reaching a complete image (100\%). At each iteration, the model predicts all masked tokens in parallel and randomly unmasks a subset. The percentages show the progress through generation steps. Generation takes hundreds of iterations regardless of feature dimensionality, making high-dimensional discrete generation computationally feasible. The visualization demonstrates a coarse-to-fine generation process, where early iterations establish overall structure and later iterations refine details.}
  \label{fig:infer}
   \vspace{-10pt}
\end{figure*}

The discretization process, although preserving continuous-level quality, yields $h \times w \times d$ discrete tokens. For example, it takes 196,608 tokens for a typical 16×16×768 configuration. The real challenge lies in how to model this massive token space: direct autoregressive generation would require $O(hwd)$ steps, while naive parallel methods fail to capture the complex dependencies within this structured tensor.

\vspace{3pt}
\noindent\textbf{Masking Across Spatial and Dimensional Axes.}
In this paper, we propose Cubic Discrete Diffusion (CubiD), which follows the discrete diffusion paradigm by treating generation as iterative denoising of masked tokens. Unlike traditional discrete diffusion methods like MaskGIT~\cite{chang2022maskgit} that mask entire spatial positions, \Ours performs fine-grained masking at the dimension level—treating the $h \times w \times d$ tensor as a unified modeling space where any subset of dimensions can be masked and predicted from the remaining visible context. This enables the model to capture rich dependencies both within and across spatial locations.

Given discrete tokens $\mathbf{q} \in \{0,...,L-1\}^{h \times w \times d}$ from dimension-wise quantization, \Ours learns to predict randomly masked tokens from visible ones. As illustrated in Figure~\ref{fig:train}(b), during training, we apply a binary mask $\mathbf{M} \in \{0,1\}^{h \times w \times d}$ where each element is independently and randomly masked. We first sample a masking ratio $r$ from a truncated Gaussian distribution:
\begin{equation}
    r \sim \text{TruncNorm}(\mu=1.0, \sigma, [0, 1.0])
    \label{eq:truncnorm}
\end{equation}
where $\mu=1.0$ is the mean and $\sigma$ is the standard deviation, with the distribution truncated to the range [0, 1]. Then, we randomly select $\lfloor r \times h \times w \times d \rfloor$ positions to mask across the entire tensor. This distribution covers the full range [0, 1] to ensure consistency with inference, which progresses from fully masked to fully unmasked. With $\mu=1.0$, it biases toward aggressive masking, encouraging the model to learn robust predictions from minimal context. Masked positions are replaced with a learnable \texttt{[MASK]} token, and the model is trained to predict the original discrete token categories at these positions through cross-entropy loss:
\begin{equation}
    \mathcal{L} = -\mathbb{E}_{\mathbf{q},\mathbf{M}} \left[ \sum_{i \in \mathbf{M}} \log p(q_i | \mathbf{q}_{\bar{\mathbf{M}}}) \right]
\end{equation}
where $\mathbf{q}_{\bar{\mathbf{M}}}$ denotes the visible tokens that provide context for prediction.

This fine-grained masking allows the model to observe partial dimensions at each location, learning how different dimensions jointly encode information and constrain each other's values. Through bidirectional attention over the partially masked tensor, the model discovers complex dependency patterns both within and across spatial positions without being constrained to predefined factorization orders.

\vspace{3pt}
\noindent\textbf{Inference.}
During inference, \Ours generates images through iterative refinement starting from a fully masked tensor. As illustrated in Figure~\ref{fig:infer}, the model begins with all tokens masked (0\%) and progressively unmasks them until reaching a complete image (100\%). At each iteration $t$, the model predicts all masked tokens simultaneously and unmasks a subset randomly. Motivated by MaskGIT~\cite{chang2022maskgit}, the number of tokens to unmask follows a cosine schedule. The schedule ensures a coarse-to-fine generation process where early iterations establish overall structure and later iterations refine details. Crucially, the parallel nature of our approach means generation requires only $O(T)$ iterations—typically hundreds of steps—regardless of the tensor dimensionality $d$, making high-dimensional discrete generation computationally feasible.

\noindent\textbf{Model Architecture.}
\Ours employs a standard Transformer architecture with bidirectional attention. As shown in Figure~\ref{fig:train}(b), each spatial position, comprising $d$ tokens, is treated as a single token for the transformer model, thereby preserving the spatial structure while enabling fine-grained predictions. Specifically, for each spatial position, we dequantize its $d$ discrete tokens back to continuous scalars (with \texttt{[MASK]} tokens mapped to a learnable value) and concatenate them into a $d$-dimensional feature vector. This results in a sequence of $h \times w$ tokens, each with dimensionality $d$.
The Transformer processes this sequence through bidirectional attention, with the sequence length remaining fixed at $h \times w$ regardless of feature dimensionality. Each output token from the Transformer is passed through an MLP-based prediction head that produces $d \times L$ logits, enabling simultaneous prediction of all $d$ dimensions at that spatial position. This design decouples computational complexity from feature dimensionality—the Transformer's sequence length depends only on spatial resolution, not on $d$.

\section{Experiments}
\label{sec:exp}

\subsection{Implementation Details}
\label{sec:implementation}

\noindent \textbf{Representation Encoders.}
We use frozen DINOv2-B~\cite{oquab2023dinov2} and SigLIP2-B~\cite{tschannen2025siglip2} as representation encoders, both producing 16×16×768 feature maps. DINOv2-B processes 224×224 images while SigLIP2-B takes 256×256 inputs. For reconstruction, we adopt decoders from~\cite{rae} that decode 256×256 images. Unless otherwise specified, we use DINOv2-B as our default encoder.

\vspace{3pt}
\noindent \textbf{Model Configurations.}
We evaluate three model sizes as shown in Table~\ref{tab:model_config}. All models use 16 attention heads with MLP ratio of 4. Unless otherwise specified, we report results using \Ours-L.
\begin{table}[t]
\centering
\small
\setlength{\tabcolsep}{3pt}
\caption{\textbf{Model sizes and architecture configurations of \Ours.}}
\vspace{-7pt}
\label{tab:model_config}
\begin{tabular}{lccc}
\toprule
Model & Hidden Dim & Blocks & Parameters \\
\midrule
\Ours-L & 1536 & 32 & 946M \\
\Ours-XL & 1920 & 32 & 1.4B \\
\Ours-XXL & 3072 & 32 & 3.7B \\
\bottomrule
\end{tabular}
\vspace{-14pt}
\end{table}

\begin{table}[t]
\centering
\small
\setlength{\tabcolsep}{3pt}
\caption{\textbf{Effect of quantization levels on reconstruction quality.} Both encoders achieve continuous-level performance with appropriate quantization levels (L=8 for DINOv2, L=16 for SigLIP2).}
\vspace{-7pt}
\label{tab:reconstruction}
\begin{subtable}[t]{0.45\linewidth}
\centering
\setlength{\tabcolsep}{1pt}
\begin{tabular}{lccc}
\toprule
DINOv2~\cite{oquab2023dinov2} & $L$ & rFID$\downarrow$ & IS$\uparrow$ \\
\midrule
Continuous & - & 0.57 & 226.9 \\
\midrule
Discrete & 2 & 1.38 & 206.1 \\
 & 4 & 0.70 & 221.1 \\
\rowcolor{gray!20} & 8 & \textbf{0.57} & 226.8 \\
 & 16 & 0.57 & \textbf{226.9} \\
\bottomrule
\end{tabular}
\caption{\textbf{DINOv2 encoder.}}
\label{tab:reconstruction_dinov2}
\end{subtable}
\hfill
\begin{subtable}[t]{0.45\linewidth}
\centering
\setlength{\tabcolsep}{1pt}
\begin{tabular}{lccc}
\toprule
SigLIP2~\cite{tschannen2025siglip2} & $L$ & rFID$\downarrow$ & IS$\uparrow$ \\
\midrule
Continuous & - & 0.69 & 217.5 \\
\midrule
Discrete & 4 & 1.54 & 193.8 \\
 & 8 & 0.92 & 210.7 \\
\rowcolor{gray!20} & 16 & \textbf{0.69} & 216.2 \\
 & 32 & 0.69 & \textbf{217.5} \\
\bottomrule
\end{tabular}
\caption{\textbf{SigLIP2 encoder.}}
\label{tab:reconstruction_siglip}
\end{subtable}
\vspace{-10pt}
\end{table}

\begin{table}[ht]
\centering
\small
\setlength{\tabcolsep}{3pt}
\caption{\textbf{Understanding performance on LLaVA benchmarks with different quantization methods.} Evaluation using SigLIP2 features. VQ: vector quantization, DQ: dimension-wise quantization. DQ maintains continuous-level performance while VQ shows significant degradation.}
\vspace{-7pt}
\label{tab:understanding}
\begin{tabular}{lccccc}
\toprule
Tokenizer & Type & GQA & TextVQA & POPE & MME \\
\midrule
SigLIP2 & Continuous & 63.2 & 59.6 & 85.4 & 1484 \\
SigLIP2-VQ & Discrete & 54.9 & 45.6 & 81.2 & 1189 \\
SigLIP2-DQ & Discrete & 63.1 & 59.8 & 85.0 & 1480 \\
\bottomrule
\end{tabular}
\vspace{-14pt}
\end{table}

\vspace{3pt}
\noindent \textbf{Training and Inference.}
Models are trained on ImageNet~\cite{imagenet} at 256×256 resolution. We use AdamW optimizer with learning rate $5 \times 10^{-5}$, cosine schedule, and 0.05 weight decay. Gradient clipping is applied at norm 3.0. Ablation studies use 150 epochs while final results are reported at 800 epochs. Generation employs iterative unmasking with cosine scheduling for mask ratios, using $T=256$ steps for ablation studies.

\vspace{3pt}
\noindent \textbf{Evaluation Metrics.}
We evaluate generation quality using Fréchet Inception Distance (FID)~\cite{fid} and Inception Score (IS)~\cite{inception_score} on ImageNet 256×256. Precision and Recall metrics~\cite{precision_recall} are reported as additional references for sample quality and diversity.

\subsection{Studies of Discretization}

In this section, we study the effects of dimension-wise quantization on high-dimensional features through reconstruction and understanding experiments.

\vspace{3pt}
\noindent\textbf{Reconstruction Quality.}
We evaluate dimension-wise quantization on two representation encoders, DINOv2-B~\cite{oquab2023dinov2} and SigLIP2-B~\cite{tschannen2025siglip2}, using their continuous reconstruction results as baselines. As shown in Table~\ref{tab:reconstruction}, discretized tokens can preserve the original continuous performance with appropriate quantization levels. Specifically, DINOv2-B achieves baseline rFID (0.57) at $L=8$, while SigLIP2-B reaches its baseline (rFID=0.69) at $L=16$. We adopt these settings for all subsequent experiments. The different optimal quantization levels likely reflect distinct feature distributions between encoders.

\vspace{3pt}
\noindent\textbf{Understanding Quality.}
To validate whether discrete tokens maintain the understanding capabilities of continuous representations, we evaluate the discrete token features on multimodal understanding tasks. 
We adopt the classic LLaVA~\cite{liu2024llavanext} framework and select SigLIP2\cite{tschannen2025siglip2} as the vision encoder for its strong cross-modal alignment. In our setup, we only replace the vision encoder features while keeping all other components unchanged. We compare three variants: (1) original continuous SigLIP2 features, (2) vector quantization~\cite{vqvae} (SigLIP2-VQ), and (3) dimension-wise quantization (SigLIP2-DQ). 
For the discrete variants, we use their dequantized features as input to LLaVA. We follow the LLaVA training protocol and evaluate on four standard benchmarks: GQA~\cite{gqa}, TextVQA~\cite{textvqa}, POPE~\cite{pope}, and MME~\cite{mme}. 
As shown in Table~\ref{tab:understanding}, SigLIP2-DQ achieves nearly identical performance to continuous features (63.1 vs 63.2 on GQA, 59.8 vs 59.6 on TextVQA), while SigLIP2-VQ shows significant degradation across all metrics. These results confirm that dimension-wise quantization preserves the semantic understanding capabilities essential for multimodal tasks.

\subsection{Studies of \Ours}

\begin{table}[t]
\centering
\small
\caption{\textbf{Ablation studies on \Ours design choices.} Gray rows indicate best results.}
\setlength{\tabcolsep}{5pt}
\vspace{-7pt}
\begin{subtable}[t]{0.45\linewidth}
\centering
\begin{tabular}{cc}
\toprule
$\sigma$ & gFID$\downarrow$ \\
\midrule
0.05 & 7.65\\
\rowcolor{gray!20} 0.10 & \textbf{5.33}\\
0.15 & 5.81\\
\bottomrule
\end{tabular}
\caption{\textbf{Mask ratio distribution.} Effect of standard deviation $\sigma$ in sampling mask ratios. Smaller $\sigma$ biases toward aggressive masking, larger $\sigma$ provides uniform coverage.}
\vspace{-3pt}
\label{tab:mask_distribution}
\end{subtable}
\hfill
\begin{subtable}[t]{0.45\linewidth}
\centering
\begin{tabular}{lc}
\toprule
Masking Strategy & gFID$\downarrow$ \\
\midrule
Per-dim & 120.03 \\
Per-spatial & 22.22\\
\rowcolor{gray!20} Per-element (Ours) & \textbf{5.33}\\
\bottomrule
\end{tabular}
\caption{\textbf{Masking granularity.} Per-dim: mask all spatial positions per dimension. Per-spatial: mask all dimensions per position. Per-element: mask independently across all axes.}
\vspace{-3pt}
\label{tab:mask_pattern}
\end{subtable}
\begin{subtable}[t]{0.45\linewidth}
\centering
\begin{tabular}{lc}
\toprule
Mask Value & gFID$\downarrow$ \\
\midrule
Fixed & 5.56\\
Random & 56.38 \\
\rowcolor{gray!20} Learned & \textbf{5.33}\\
\bottomrule
\end{tabular}
\caption{\textbf{Mask value.} Fixed, random, or learned mask token.}
\vspace{-3pt}
\label{tab:mask_value}
\end{subtable}
\hfill
\begin{subtable}[t]{0.45\linewidth}
\centering
\begin{tabular}{cc}
\toprule
Steps & gFID$\downarrow$ \\
\midrule
64 & 9.14\\
256 & 5.33\\
\rowcolor{gray!20} 512 & \textbf{5.25}\\
1024 & 5.25\\
\bottomrule
\end{tabular}
\caption{\textbf{Inference steps.} Effect of inference steps $T$.}
\vspace{-3pt}
\label{tab:num_steps}
\end{subtable}
\begin{subtable}[t]{0.45\linewidth}
\centering
\begin{tabular}{cc}
\toprule
Params & gFID$\downarrow$ \\
\midrule
946M & 5.25 \\
1.4B & 4.91\\
\rowcolor{gray!20} 3.7B & \textbf{4.68}\\
\bottomrule
\end{tabular}
\caption{\textbf{Model scaling.} Effect of model size.}
\vspace{-3pt}
\label{tab:model_scaling}
\end{subtable}
\hfill
\begin{subtable}[t]{0.45\linewidth}
\centering
\begin{tabular}{lc}
\toprule
Encoder & gFID$\downarrow$ \\
\midrule
\rowcolor{gray!20} DINOv2 & \textbf{5.25}\\
SigLIP2 & 5.87\\
\bottomrule
\end{tabular}
\caption{\textbf{Representation encoder.} DINOv2 vs. SigLIP2.}
\label{tab:encoder}
\end{subtable}
\vspace{-15pt}
\label{tab:ablations}
\end{table}

\noindent\textbf{Mask Ratio Distribution.}
Masking is the core operation of our discrete diffusion approach, and the distribution of masking ratios critically affects what patterns the model learns. We sample the masking ratio $r$ from a truncated Gaussian distribution (Eq.~\ref{eq:truncnorm}) with $\mu=1.0$ and varying standard deviation $\sigma$. The parameter $\sigma$ controls the diversity of masking scenarios: small $\sigma$ concentrates sampling around high masking ratios, forcing the model to learn from minimal context, while larger $\sigma$ provides more uniform coverage across the [0, 1] range. Table~\ref{tab:mask_distribution} shows that $\sigma=0.10$ achieves optimal performance (gFID=5.33). Too small values ($\sigma=0.05$) degrade generation quality—the model overfits to heavily masked patterns without learning the full distribution. This optimal setting suggests that high-dimensional features benefit from aggressive masking during training, likely due to their inherent redundancy.

\begin{figure}[!t]
\centering
\includegraphics[width=\linewidth]{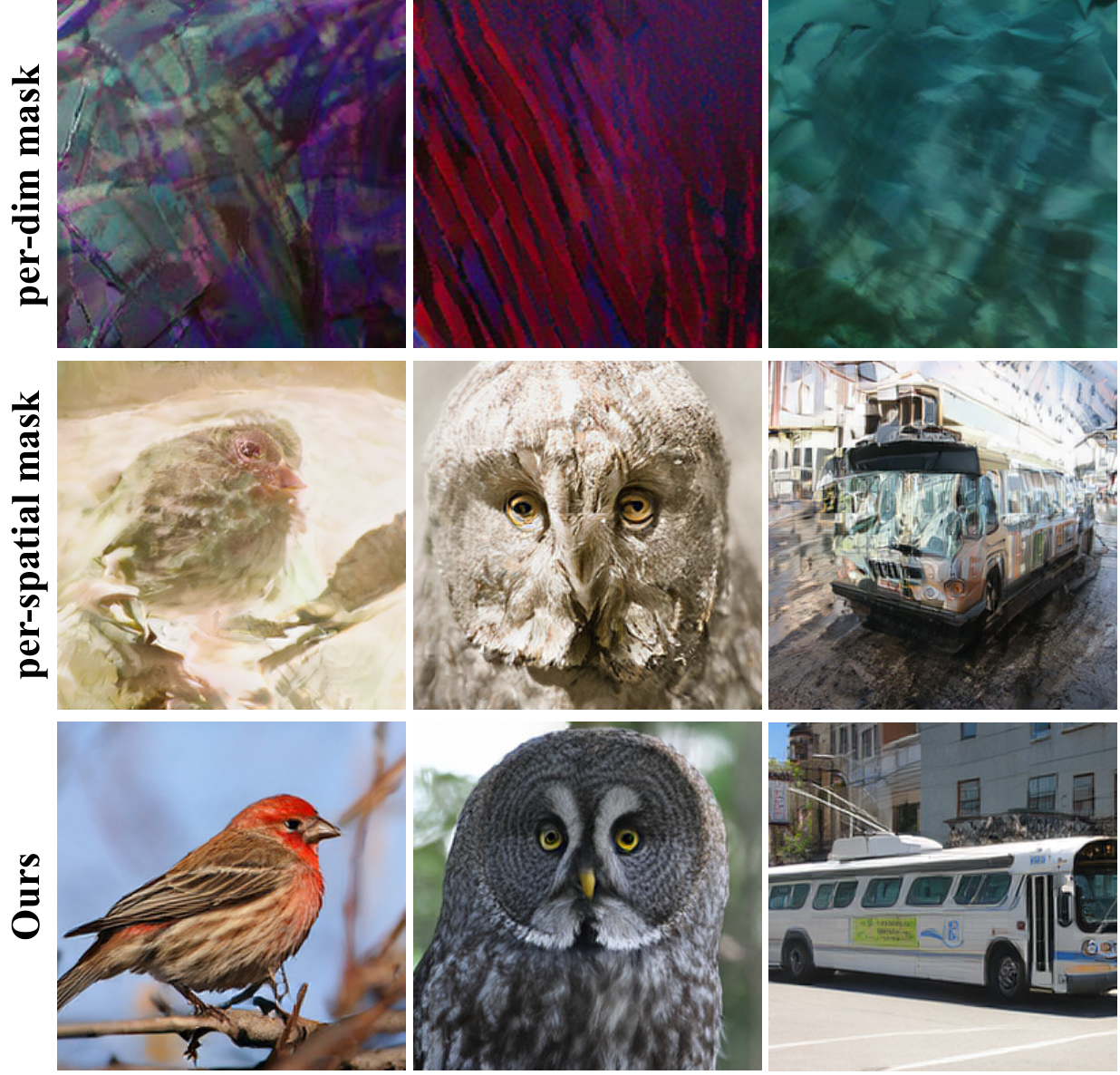}
\vspace{-2.2em} 
\caption{\textbf{Qualitative comparison of different masking strategies.} Top row: Per-dim masking completely fails, producing severe texture-like artifacts. Middle row: Per-spatial masking generates images with significant local inconsistencies and blurry details. Bottom row: Our per-element masking produces clear, coherent images with fine details. The dramatic quality difference validates that high-dimensional tokens require fine-grained masking across both spatial and dimensional axes.}
\vspace{-15pt}
\label{fig:mask_comp}
\end{figure}

\vspace{3pt}
\noindent\textbf{Masking Strategy.}
We investigate different masking strategies for the $h \times w \times d$ representation tensor. Table~\ref{tab:mask_pattern} and Figure~\ref{fig:mask_comp} compare three approaches: (1) Per-dim masking, where all spatial positions for each dimension are masked together; (2) Per-spatial masking, where all dimensions at each spatial position are masked together; and (3) Per-element masking, our approach that independently masks individual elements across the tensor. The results show obvious performance differences: per-dim masking completely fails (gFID=120.03) with severe texture artifacts, while per-spatial masking produces blurry, locally inconsistent images (gFID=22.22). In contrast, our per-element masking achieves strong performance (gFID=5.33). This is because elements within the same spatial location or dimension exhibit strong dependencies and cannot be treated as independent units for parallel sampling. The 768 dimensions at each spatial position jointly encode semantic information—masking them together (per-spatial) prevents the model from leveraging these within-position correlations. Per-dim masking performs even worse as it requires all spatial positions to be generated in parallel, destroying spatial coherence entirely. Our per-element masking enables the model to observe partial information along both axes during training and generation, utilizing bidirectional attention to capture dependencies across the tensor. This validates the necessity of fine-grained masking for modeling high-dimensional discrete tokens, where neither spatial positions nor dimensions can be fully decoupled.

\vspace{3pt}
\noindent\textbf{Mask Token Design.}
We investigate different strategies for the mask token value used during training and inference. Table~\ref{tab:mask_value} compares three approaches: (1) Fixed: using a constant value (zero in our experiments), (2) Random: sampling from the discrete codebook at each masking operation, and (3) Learned: treating the mask token as a learnable parameter. The learned mask token achieves the best performance (gFID=5.33), while random sampling performs poorly (gFID=56.38). The failure of random sampling likely stems from the model's inability to distinguish between actual content tokens and randomly sampled mask tokens, as both come from the same codebook distribution. In contrast, a learned mask token can evolve during training to be maximally distinguishable from content tokens, facilitating more effective learning.

\begin{table*}[t]
\centering
\small
\resizebox{.85\textwidth}{!}{%
\begin{tabular}{llc@{\hspace{10pt}}cccc@{\hspace{10pt}}cccc}
\toprule
\multirow{2}{*}{Method} & \multirow{2}{*}{Latent Dim} & \multirow{2}{*}{\#Params} & \multicolumn{4}{c}{Generation@256 w/o guidance} & \multicolumn{4}{c}{Generation@256 w/ cfg or re} \\
\cmidrule(lr){4-7} \cmidrule(lr){8-11}
& & & gFID$\downarrow$ & IS$\uparrow$ & Prec.$\uparrow$ & Rec.$\uparrow$ & gFID$\downarrow$ & IS$\uparrow$ & Prec.$\uparrow$ & Rec.$\uparrow$ \\
\midrule
\multicolumn{11}{l}{\textbf{\textit{Discrete Diffusion Models with Low-dimensional Tokens}}} \\
MaskGIT~\cite{chang2022maskgit} & 16 & 227M & 6.18 & 182.1 & 0.80 & 0.51 & 4.02$^\text{re}$ & 355.6$^\text{re}$ & - & - \\
VQ-Diffusion~\cite{vq-diffusion} &16&370M & 11.89&-&-&-&-&-&-&-\\
Token-Critic~\cite{token_critic} & 16 & 368M & 4.69 & 174.5 & 0.76 & 0.51 & - & - & - & - \\

DPC~\cite{DPC} & 16 & 454M & 4.45 & 244.8 & - & - & - & - & - & - \\

TiTok-S-128~\cite{titok} & 16 & 287M & - & - & - & - & 1.97 & 281.8 & - & - \\
\midrule
\multicolumn{11}{l}{\textbf{\textit{Discrete Autoregressive Models with Low-dimensional Tokens}}} \\
VQVAE2~\cite{vqvae2} & 64 & 13.5B & 31.11 & 45 & 0.36 & 0.57 & - & - & - & - \\
VQGAN~\cite{esser2020taming} & 128 & 1.4B & 15.78 & 74.3 & - & - & 5.20$^\text{re}$ & 280.3$^\text{re}$ & - & - \\
ViT-VQGAN~\cite{vit-vqgan} & 32 & 1.7B & 4.17 & 175.1 & - & - & 3.04$^\text{re}$ & 227.4$^\text{re}$ & - & - \\
RQTran.~\cite{rq} & 16 & 3.8B & 7.55 & 134.0 & - & - & 3.80$^\text{re}$ & 323.7$^\text{re}$ & - & - \\
LlamaGen-XXL~\cite{sun2024autoregressive} & 8 & 1.4B & 14.6 & 86.3 & 0.63 & 0.68 & 2.34 & 253.9 & 0.81 & 0.60 \\
VAR~\cite{tian2024visual} & 32 & 2.0B & 2.16 & 288.7 & 0.81 & 0.61 & 1.97 & 334.7 & 0.81 & 0.61 \\
VFMTok-XXL ~\cite{zheng2025vision-foundation} & 12 & 1.4B & 1.95 & 259.3 & 0.82 & 0.62 & 2.19 & 278.0 & 0.83 & 0.60 \\
VFMTok-3B ~\cite{zheng2025vision-foundation} & 12 & 3.1B & 2.04 & 267.6 & 0.82 & 0.61 & 2.07 & 280.4 & 0.81 & 0.62 \\
\midrule
\multicolumn{11}{l}{\textbf{\textit{Discrete Models with High-dimensional Tokens}}} \\
\rowcolor{gray!20} CubiD-L & 768 & 946M & 2.38 & 213.1 & 0.84 & 0.57 & 2.37 & 213.4 & 0.84 & 0.57 \\
\rowcolor{gray!20} CubiD-XL & 768 & 1.4B & 2.06 & 216.4 & 0.83 & 0.58 & 2.04 & 217.0 & 0.83 & 0.59 \\
\rowcolor{gray!20} CubiD-XXL & 768 & 3.7B & 2.02 & 214.8 & 0.81 & 0.61 & 1.88 & 247.0 & 0.83 & 0.58 \\
\bottomrule
\end{tabular}
}
\vspace{-7pt}
\caption{\textbf{Discrete generation methods on ImageNet~\cite{imagenet} 256×256.} Latent Dim denotes the original dimensionality of the latent space (features before vector quantization for low-dimensional methods, before and after dimension-wise quantization for CubiD). Results with superscript "re" denote rejection sampling. CubiD is the first and only discrete method to directly generate with native high-dimensional representation tokens (768d), while all other methods use compressed or low-dimensional tokens (mostly below 32).}
\label{tab:main_results}
\vspace{-14pt}
\end{table*}

\vspace{3pt}
\noindent\textbf{Number of Iterations.}
Table~\ref{tab:num_steps} illustrates the effect of inference steps on generation quality. In this experiment with DINOv2, our model needs to generate $h \times w \times d = 16 \times 16 \times 768 = 196,608$ discrete tokens for each image. Despite this massive token count, our method requires only hundreds of iterations to achieve high-quality generation. Performance improves from 64 to 256 steps and saturates around 512 iterations (gFID=5.25). This is remarkably efficient compared to autoregressive methods that would require all 196,608 sequential steps.

\vspace{3pt}
\noindent\textbf{Model Scaling.}
Table~\ref{tab:model_scaling} shows results for models ranging from 946M to 3.7B parameters. We observe consistent improvement in generation quality as model size increases, with gFID decreasing from 5.25 for the 946M model to 4.68 for the 3.7B model. This scaling behavior demonstrates that our cubic discrete formulation effectively leverages increased model capacity, exhibiting strong scaling properties similar to other discrete generative models like autoregressive models. The steady improvement across model sizes suggests that our method can benefit from further scaling, making it a promising direction for high-quality representation-based image generation at larger scales.

\vspace{3pt}
\noindent\textbf{Representation Encoder.}
We evaluate \Ours with different representation encoders to assess generalization. Table~\ref{tab:encoder} compares DINOv2~\cite{oquab2023dinov2} and SigLIP2~\cite{tschannen2025siglip2} encoders, both producing 16×16×768 feature maps. Both encoders work well with our generation model, achieving gFID scores of 5.25 and 5.87 respectively with limited epochs. DINOv2 achieves slightly better generation quality, likely due to its ImageNet pretraining being better aligned with ImageNet-based evaluation metrics.
The consistent performance across both encoders, despite their different training objectives, demonstrates the robustness of our approach.

\subsection{Main Results}

Table~\ref{tab:main_results} presents our main results on ImageNet 256×256 class-conditional generation, comparing CubiD with existing discrete generation methods. We organize methods into three categories: discrete diffusion with low-dimensional tokens, discrete autoregressive with low-dimensional tokens, and discrete models with high-dimensional tokens.
CubiD is the only method that directly generates with native high-dimensional representation tokens. All existing methods operate in latent spaces ranging from 8 to 128 dimensions, with most below 32. Despite the increased complexity of modeling high-dimensional tokens, CubiD-XXL achieves state-of-the-art discrete generation with a gFID of 1.88. Notably, representation tokens show reduced dependency on classifier-free guidance—even without guidance, CubiD-XXL achieves 2.02 gFID, outperforming most VAE-based methods without guidance (e.g., MaskGIT at 6.18 and LlamaGen-XXL at 14.6).
While VFMTok also leverages representation features, it introduces deformable attention and region-adaptive mechanisms to reorganize the original features into 12-dimensional VQ tokens. This reorganization enables tractable autoregressive generation but fundamentally alters the token space, potentially limiting their use for understanding tasks. Moreover, VFMTok shows limited scaling benefits—performance slightly degrades from VFMTok-XXL (1.95 gFID) to VFMTok-3B (2.04 gFID). In contrast, CubiD demonstrates consistent improvement with scale, from 2.37 (L) to 2.04 (XL) to 1.88 (XXL) with cfg, while generating directly in the original high-dimensional representation space without any reorganization or compression. These results illustrate the effectiveness of our discrete diffusion approach for high-dimensional token generation.

\section{Conclusion}
\label{sec:conclusion}

In this work, we introduce CubiD, a novel discrete generative model that directly models native high-dimensional representation tokens for the first time. We achieve this through fine-grained masking across the entire spatial-dimensional tensor, transforming the intractable problem of generating hundreds of thousands of sequential tokens into manageable parallel iterations. Our work demonstrates that discrete generation with standard cross-entropy loss can achieve state-of-the-art results even in the challenging regime of high-dimensional tokens, without requiring compression or reorganization of the original representation space. The preservation of native representation ability enables the same discrete tokens to serve both understanding and generation tasks, eliminating the need for separate tokenization schemes across tasks. We hope our work will inspire future research on unified multimodal architectures.

\section*{Acknowledgment}
This work is supported in part by the Research Grant Council of Hong Kong through the NSFC-RGC Joint Research Scheme under grant N\_HKU769/25.
The authors are grateful to Boyang Zheng for helpful discussions on RAE and to Difan Zou, Yi Zhang, Yujin Han and Yuanzhi Zhu for valuable feedback on the early version of this work.

{
    \small
    \bibliographystyle{ieeenat_fullname}
    \bibliography{main}
}

\clearpage
\setcounter{page}{1}
\setcounter{section}{0}
\renewcommand{\thesection}{\Alph{section}}
\maketitlesupplementary

\section*{Appendix}
\addcontentsline{toc}{section}{Appendix}

The supplementary material includes the following additional information:
\begin{itemize}
    \item Sec.~\ref{sec:A} provides more implementation details for generation and understanding experiments.
    \item Sec.~\ref{sec:B} presents additional experiments of CubiD on low-dimensional tokens.
    \item Sec.~\ref{sec:C} discusses limitations.
    \item Sec.~\ref{sec:D} showcases additional image generation results.
\end{itemize}

\section{Implementation Details}
\label{sec:A}

\subsection{Generation Experiments}

\textbf{Additional Training Details.} 
We train all CubiD models on the ImageNet-1K~\cite{imagenet} training set, consisting of 1,281,167 images across 1,000 object classes. Beyond the details provided in the main paper, we use a batch size of 2048 distributed across all GPUs. We employ mixed precision training with fp16 to reduce memory consumption and accelerate training. An exponential moving average (EMA) of model weights is maintained with momentum 0.9999 for stable evaluation. The learning rate warmup is applied for the first 100 epochs. We adopt the noise-augmented decoder from~\cite{rae}, which injects Gaussian noise into clean latents during decoder training to improve robustness to imperfect generative outputs.

\subsection{Understanding Experiments}

\textbf{Training.} 
To validate the understanding performance of our discretized tokens, we adopt the classic LLaVA~\cite{liu2024llavanext} visual instruction tuning framework, and perform experiments with the original representations and discretized tokens. Following its standard protocol, we first perform pretrain on 558K LAION-CC-SBU~\cite{li2022blip} subset for 1 epoch, then conduct visual instruction tuning on the LLaVA-Instruct-665K dataset for 1 epoch. We use Vicuna-13B-v1.5~\cite{zheng2023judging} as the language backbone and maintain all original hyperparameters, with the only modification being the replacement of continuous vision features with their quantized counterparts.

\textbf{Evaluation.} 
We evaluate on four standard benchmarks from the LLaVA evaluation suite: GQA~\cite{gqa} for compositional visual reasoning, TextVQA~\cite{textvqa} for text recognition and understanding in images, POPE~\cite{pope} for assessing hallucination tendencies, and MME~\cite{mme} for comprehensive multimodal perception capabilities. These benchmarks collectively measure whether the quantized representations maintain the diverse understanding abilities required for multimodal tasks.

\section{CubiD on Low-Dimensional Tokens}
\label{sec:B}

\begin{table}[t]
\centering
\small
\caption{\textbf{CubiD on low-dimensional tokens on ImageNet 512×512.} Results using DC-AE-f32c32 tokenizer producing 32-dimensional tokens.}
\vspace{-7pt}
\label{tab:low_dim}
\begin{tabular}{lccc}
\toprule
Method & Params (B) & gFID↓ & IS↑ \\
\midrule
SiT-XL~\cite{ma2024sit} & 0.67 & 2.41 & 131.37 \\
USiT-H~\cite{dcae} & 0.50 & 1.89 & 174.58 \\
USiT-2B~\cite{dcae} & 1.58 & 1.72 & 187.68 \\
\midrule
\rowcolor{gray!20} \Ours & 0.95 & \textbf{1.58} & \textbf{188.70} \\
\bottomrule
\vspace{-7pt}
\end{tabular}
\end{table}

To validate the generality of CubiD beyond high-dimensional representations, we conduct experiments on traditional low-dimensional tokens. 

\subsection{Traditional Reconstruction-based Tokens}

We employ DC-AE-f32c32~\cite{dcae}, a state-of-the-art autoencoder with patch size 32 that produces 32-dimensional tokens. For 512×512 images, this results in 16×16×32 discrete tokens after dimension-wise quantization, which are significantly more compact than the 32×32×768 tokens in our main experiments.
As shown in Table~\ref{tab:low_dim}, CubiD achieves 1.58 gFID and 188.7 IS on ImageNet 512×512, outperforming previous state-of-the-art methods using the same tokenizer, including USiT-2B (1.72 gFID) despite using fewer parameters. This demonstrates that our cubic discrete diffusion formulation is effective across different token dimensionalities.

\subsection{Compressed Representation Tokens}

\begin{table}[t]
\centering
\small
\caption{\textbf{CubiD with compressed representation tokens on ImageNet 256×256.} Features compressed from 768d to 32d.}
\vspace{-7pt}
\label{tab:compressed}
\begin{tabular}{lccc}
\toprule
Method & Token Dim & gFID↓ & IS↑ \\
\midrule
\rowcolor{gray!20} CubiD & 32 & \textbf{1.55} & \textbf{296.5} \\
\bottomrule
\end{tabular}
\vspace{-7pt}
\end{table}

To explore the generation-understanding trade-off, we investigate CubiD's performance on compressed representation tokens. We reduce the original high-dimensional features to 32 dimensions using a learned projection layer optimized for reconstruction quality.
Table~\ref{tab:compressed} shows that compressed 32-dimensional tokens achieve strong generation performance (1.55 gFID, 296.5 IS). While lower-dimensional spaces naturally facilitate easier generation, this compression inevitably degrades the representation quality needed for understanding tasks. Therefore, we choose to model the original high-dimensional tokens to preserve both generation and understanding capabilities.

\section{Limitations}
\label{sec:C}

While CubiD demonstrates the feasibility of discrete generation on high-dimensional representation tokens, several limitations remain.

\noindent\textbf{Dependence on Representation Encoder.} Since CubiD operates on features from a frozen pretrained encoder, the reconstruction quality sets an upper bound on generation quality. In our experiments, the reconstruction PSNR is approximately 18 dB, which limits the fine-grained details in generated images. Improving the reconstruction capability of representation autoencoders remains a valuable direction for future work.

\noindent\textbf{Gap with Continuous Generation.} Although discrete generation offers advantages for unified multimodal modeling through a shared cross-entropy objective, there still exists a gap compared to continuous diffusion methods such as RAE~\cite{rae}. We believe this gap can be further narrowed with advances in discrete generative modeling.

\noindent\textbf{Inference Efficiency.} CubiD requires more generation steps than continuous diffusion models. Achieving high-quality generation typically requires hundreds to a thousand steps. Accelerating discrete diffusion inference, potentially through techniques developed for discrete language models, remains an important direction for future work.

\section{More Visualization Results}
\label{sec:D}

\begin{figure*}[h]
  \centering
  \includegraphics[width=\linewidth]{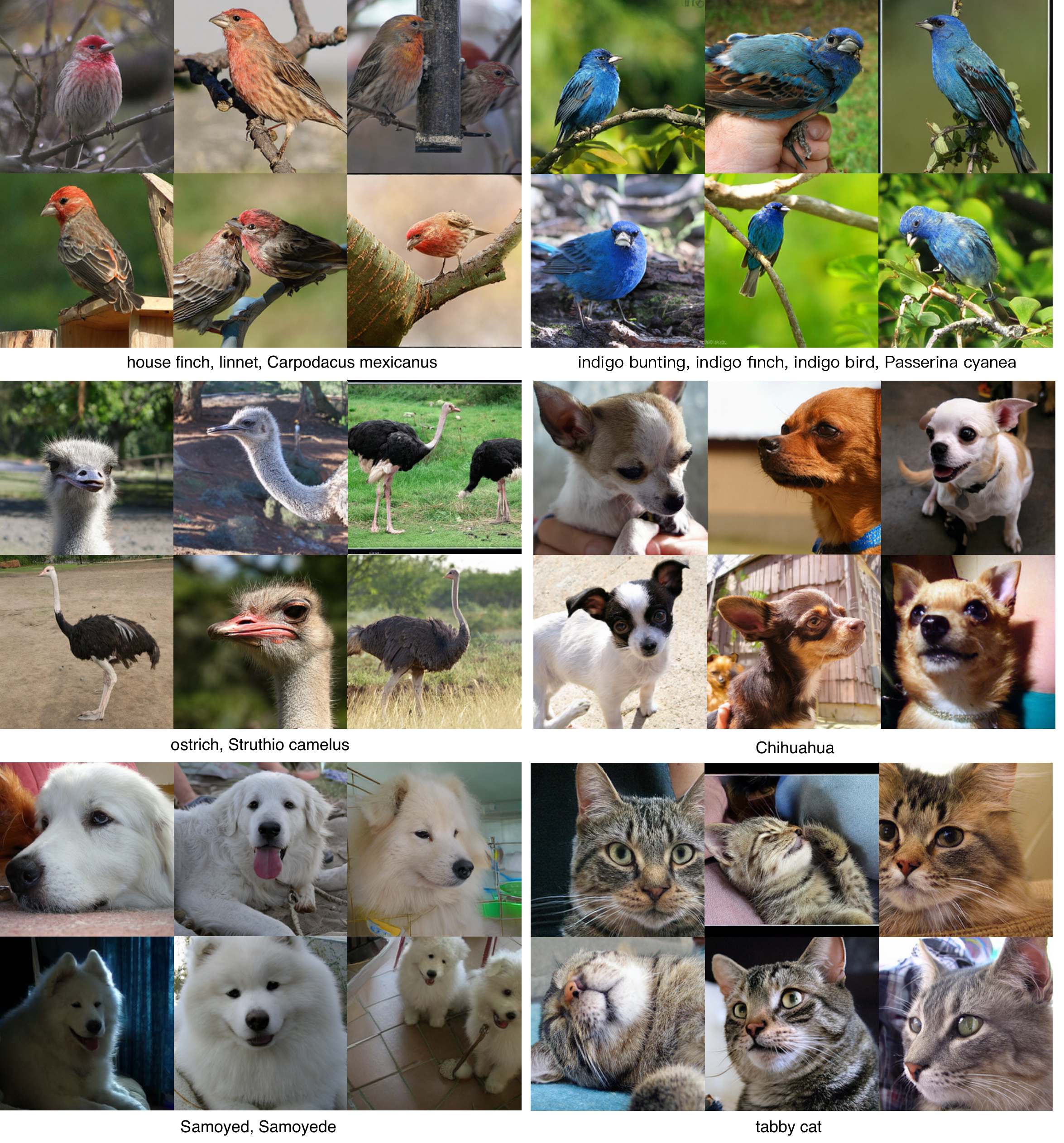}
  \caption{\textbf{Uncurated samples on ImageNet 256×256 using ~\Ours-XXL conditioned on the specified classes.}} 
  \label{fig:vis1}
\end{figure*}

\begin{figure*}[h]
  \centering
  \includegraphics[width=\linewidth]{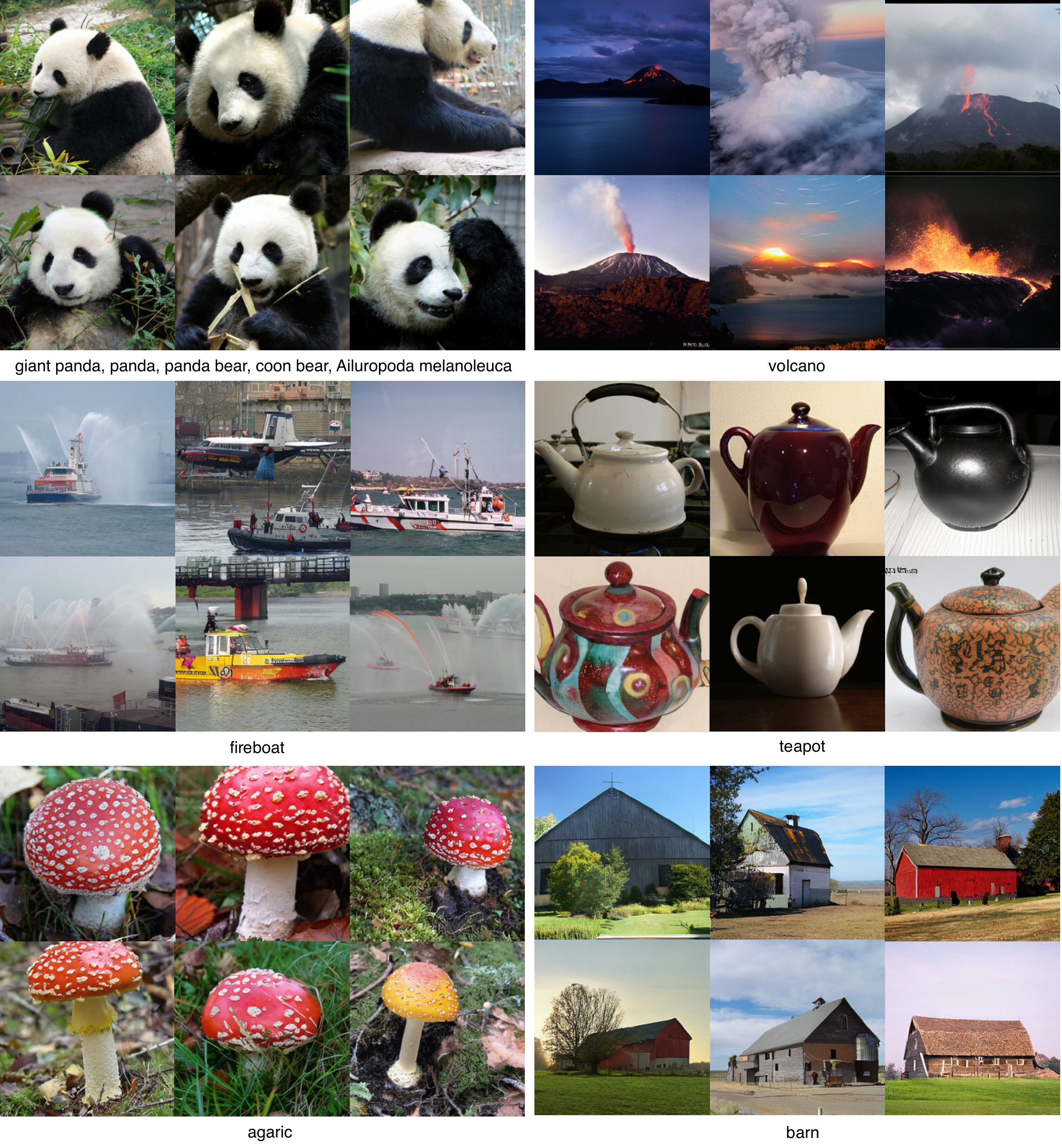}
  \caption{\textbf{Uncurated samples on ImageNet 256×256 using ~\Ours-XXL conditioned on the specified classes.}} 
  \label{fig:vis2}
\end{figure*}

\end{document}